\documentclass[conference]{IEEEtran}
\IEEEoverridecommandlockouts
\usepackage{cite}
\usepackage{amsmath,amssymb,amsfonts}
\usepackage{algorithmic}
\usepackage{graphicx}
\usepackage{textcomp}
\usepackage{xcolor}
\usepackage{url}
\usepackage{hyperref}
\usepackage{float}
\usepackage{tabularx}
\def\BibTeX{{\rm B\kern-.05em{\sc i\kern-.025em b}\kern-.08em
    T\kern-.1667em\lower.7ex\hbox{E}\kern-.125emX}}
\begin{document}

\title{A Novel Multi-branch ConvNeXt Architecture\\for Identifying Subtle Pathological Features\\in CT Scans}

\author{
\IEEEauthorblockN{Irash Perera}
\IEEEauthorblockA{
\textit{Department of Computer Science and Engineering} \\
\textit{University of Moratuwa} \\
Colombo, Sri Lanka \\
irash.21@cse.mrt.ac.lk
}
\and
\IEEEauthorblockN{Uthayasanker Thayasivam}
\IEEEauthorblockA{
\textit{Department of Computer Science and Engineering}\\
\textit{University of Moratuwa} \\
Colombo, Sri Lanka \\
rtuthaya@cse.mrt.ac.lk
}
}
\maketitle

\begin{abstract}
Intelligent analysis of medical imaging plays a crucial role in assisting clinical diagnosis, especially for identifying subtle pathological features. This paper introduces a novel multi-branch ConvNeXt architecture designed specifically for the nuanced challenges of medical image analysis. While applied here to the specific problem of COVID-19 diagnosis, the methodology offers a generalizable framework for classifying a wide range of pathologies from CT scans. The proposed model incorporates a rigorous end-to-end pipeline, from meticulous data preprocessing and augmentation to a disciplined two-phase training strategy that leverages transfer learning effectively. The architecture uniquely integrates features extracted from three parallel branches: Global Average Pooling, Global Max Pooling, and a new Attention-weighted Pooling mechanism. The model was trained and validated on a combined dataset of 2,609 CT slices derived from two distinct datasets. Experimental results demonstrate a superior performance on the validation set, achieving a final ROC-AUC of 0.9937, a validation accuracy of 0.9757, and an F1-score of 0.9825 for COVID-19 cases, outperforming all previously reported models on this dataset. These findings indicate that a modern, multi-branch architecture, coupled with careful data handling, can achieve performance comparable to or exceeding contemporary state-of-the-art models, thereby proving the efficacy of advanced deep learning techniques for robust medical diagnostics.
\end{abstract}


\section{Introduction and Background}
\begin{figure}[h]
    \centering
    \includegraphics[width=.9\linewidth]{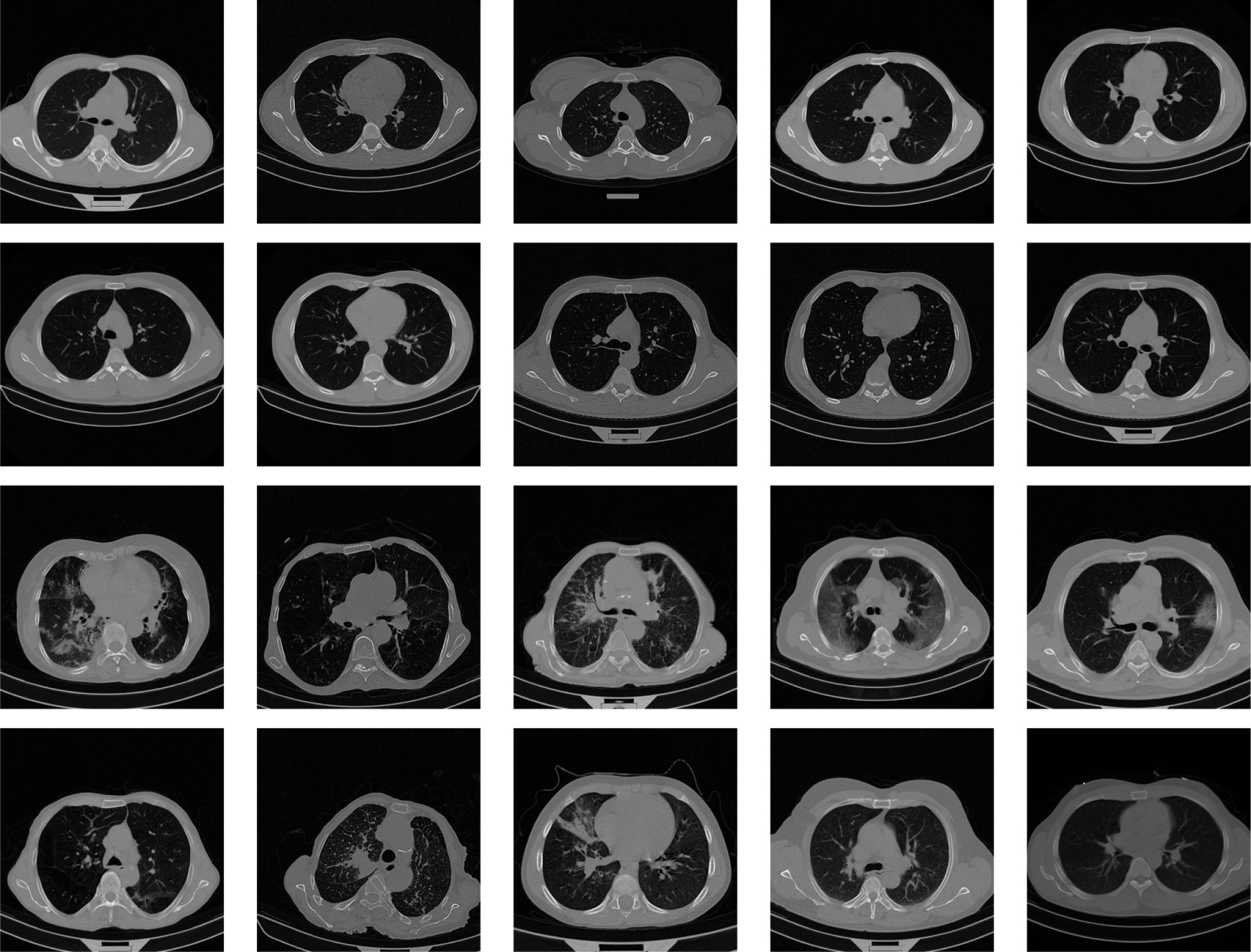}
    \caption{Examples of CT scans, first two rows contain images from healthy subjects, whereas the last two rows contain images from COVID-19 patients.\cite{b4}}
    \label{fig:covid-non-covid}
\end{figure}

The rapid global spread of Coronavirus Disease 2019 (COVID-19) highlighted the critical need for rapid and accurate diagnostic tools to manage the pandemic and mitigate its spread. While Reverse Transcription Polymerase Chain Reaction (RT-PCR) tests were the gold standard for confirmation, their shortage during peak outbreaks necessitated alternative diagnostic methods.\cite{b1} Computed Tomography (CT) scans emerged as a valuable tool for screening and diagnosing COVID-19, as they have been shown to be more sensitive than RT-PCR tests.\cite{b2} The analysis of CT scans is a time-intensive process that requires specialized medical expertise, which can be a significant bottleneck, especially for overwhelmed healthcare systems or in underdeveloped areas. As illustrated in Figure \ref{fig:covid-non-covid}, the CT scans of COVID-19 and non-COVID patients exhibit no readily apparent differences discernible without expert evaluation.

To address this challenge, artificial intelligence (AI) methods, particularly deep learning, have been developed to automate the screening of COVID-19 from CT images.\cite{b1} The development and validation of these AI models, however, were severely hampered by the scarcity of large, publicly available CT datasets due to patient privacy concerns.\cite{b3} In response, several pioneering datasets were created, including the \textbf{COVID-19 CT Lung and Infection Segmentation Dataset}\cite{b18}, which was instrumental in advancing early AI-based diagnostic research.\cite{b1} The initial models developed using these datasets, while promising, achieved performance levels that have since been surpassed by more recent state-of-the-art (SOTA) approaches.\cite{b4} This disparity in performance underscores a critical research opportunity: to re-evaluate these foundational models and enhance their capabilities using contemporary deep learning techniques to achieve clinically robust, SOTA performance.

\section{Related Work}
Research in automated diagnosis from medical imaging, particularly CT scans, has gained significant attention with the adoption of artificial intelligence and deep learning techniques.Grewal et al. \cite{b6} integrated DenseNet architecture with a recurrent neural network layer to analyze 77 brain CT scans, achieving a CT-level hemorrhage prediction accuracy of 81.82\% through their RADnet model. Similarly, Song et al. \cite{b7} developed three deep learning models, convolutional neural networks (CNNs), deep neural networks (DNNs), and stacked autoencoders (SAEs), for lung cancer classification, where the CNN outperformed the others in terms of accuracy. In another study, Gonzalez et al. \cite{b8} applied CNN-based analysis to identify and stage chronic obstructive pulmonary disease (COPD), while also predicting acute respiratory disease (ARD) occurrences and mortality risks in smokers. Early CT diagnostic models primarily employed conventional CNN backbones. For classification, transfer learning with ResNet‐ and DenseNet‐type networks proved effective.\cite{b9} These networks have been successfully leveraged to classify, segment, and detect abnormalities in various medical images, including CT scans, by learning and extracting hierarchical features from large datasets. However, the direct application of models pre-trained on natural image datasets to the medical domain proved challenging due to the inherent differences in image characteristics, such as lower contrast, noise, and artifacts, which provide limited tissue descriptions. This limitation necessitated the development of domain-specific enhancements.

During the COVID-19 outbreak, the critical need for rapid and accurate COVID-19 screening spurred a wave of research focused on automating the diagnosis from chest CT images.Key indicators observable in CT scans for COVID-19 detection include ground-glass opacities, consolidation, reticular patterns, and the crazy paving patterns.\cite{b10} Also there is a study to examine the relationship between these chest CT findings and the clinical manifestations of COVID-19 pneumonia. \cite{b11} When it comes to detect COVID-19 from CT scans, initial works served as a proof-of-concept, experimenting with well-known pre-trained CNN architectures including ResNet50, DenseNet169,EfficentNetB1 and VGG16. While these models showed the viability of deep learning for this task, their performance, while promising, often fell short of the clinical robustness required for widespread deployment. This early stage highlighted a critical research opportunity to re-evaluate and enhance these foundational models using contemporary techniques to achieve superior, state-of-the-art performance.\cite{b12}

In recent work, a Fine\_DenseNet-based deep learning model combined with an Improved Generative Adversarial Network (IGAN\_AHb) and optimized using the Artificial Hummingbird algorithm was proposed for automated multi-class COVID-19 detection from chest CT images, achieving a high classification accuracy of 95.73\%. \cite{b13}Also, recent research has leveraged pre-trained deep neural networks combined with CycleGAN-based data augmentation for automated COVID-19 detection from CT images with high accuracy, using a dataset of 3,163 images from 189 patients, while also providing interpretability through Grad-CAM visualization and calibration-based reliability assessment. \cite{b14} Another recent study proposed a hybrid model, ViTGNN, which combines Convolutional Neural Networks, Graph Neural Networks, and Vision Transformers for COVID-19 detection from CT scans, achieving high diagnostic performance with an accuracy of 95.98\%, precision of 96.07\%, recall of 96.01\%, F1-score of 95.98\%, and AUC of 98.69\%.\cite{b15} Also, a three-layer stacked multimodal framework integrating eight pre-trained transfer learning models has been proposed for deep feature extraction from large COVID-19 chest radiographic datasets, achieving high performance with an accuracy of 95.79\%, precision of 95.44\%, and recall of 96.65\%, providing an effective and computationally efficient approach for COVID-19 diagnosis.\cite{b16} Another dual-channel convolutional neural network (CNN) framework has been proposed for detecting diverse COVID-19 variants from lung CT scans, combining texture-based and spatial feature analysis with dynamic textural pattern learning, achieving high accuracy of 94.63\% and 95.47\% on the COVID-349 and Italian COVID-Set datasets, respectively, and demonstrating superior precision, recall, and diagnostic reliability compared to existing methods.\cite{b17}

\section{Method}
This section details the systematic and rigorous approach used to develop and train the proposed model. It begins with the initial steps of data acquisition and preprocessing, followed by image enhancement techniques and the extraction of the lung regions of interest. A robust data augmentation pipeline is then described, which addresses class imbalance by synthesizing new samples for the minority class. The core of the methodology is the introduction of a novel multi-branch ConvNeXt architecture designed to capture both global and fine-grained features. The section concludes with a detailed explanation of the two-phase training strategy, which effectively leverages transfer learning to adapt the model to the specific domain of medical imaging.

\subsection{Data Acquisition and Preprocessing}
For this study, two contemporary and higher-quality datasets from \cite{b5} and \cite{b18}, were utilized to build a robust training and validation set.

\begin{itemize}
    \item \textbf{COVID-19 CT Lung and Infection Segmentation Dataset}: This dataset contains 20 labeled COVID-19 CT scans. Left lung, right lung, and infections are labeled by two radiologists and verified by an experienced radiologist. \cite{b18}
    \item \textbf{MedSeg Covid Dataset 2}: This contains 9 labeled axial volumetric CTs from Radiopaedia. Both positive and negative slices (373 out of the total of 829 slices have been evaluated by a radiologist as positive and segmented). \cite{b5}
\end{itemize}

A critical methodological choice in this work was to combine these two datasets, resulting in a total of 2,609 CT slices. This synthesis of data from different sources is a common strategy in modern deep learning research to increase the sample size and improve the model's ability to generalize to variations in data source, a common challenge in medical imaging where datasets often originate from different scanners or clinical settings. The increase in sample size is a key factor in training a more robust model that is less prone to overfitting and can learn more complex patterns from the data.

The foundation of the proposed methodology is a robust data pipeline that handles the specific characteristics of CT scans. The process begins with loading the volumetric NIfTI files. This pipeline is designed to load each volumetric scan and perform an initial set of transformations. The image array is first rotated to the correct orientation, and then a subset of slices, specifically from 20\% to 80\% of the total, is selected. This step is crucial for discarding uninformative slices at the top and bottom of the scan, which typically do not contain the primary lung regions of interest. Each selected slice is then resized to a uniform dimension of 512x512 pixels and normalized to a range of 0 to 1.

\subsection{Image Enhancement and Region of Interest Extraction}
To prepare the images for classification, two key enhancement and extraction steps are applied. The first is Contrast Limited Adaptive Histogram Equalization (CLAHE). This technique improves the local contrast of the CT images, which is essential for highlighting subtle pathological features like ground-glass opacities. Unlike standard histogram equalization, CLAHE avoids over-amplifying noise in homogeneous regions by operating on small, distinct tile grids and limiting the contrast enhancement.

\begin{figure}[h]
    \centering
    \includegraphics[width=.8\linewidth]{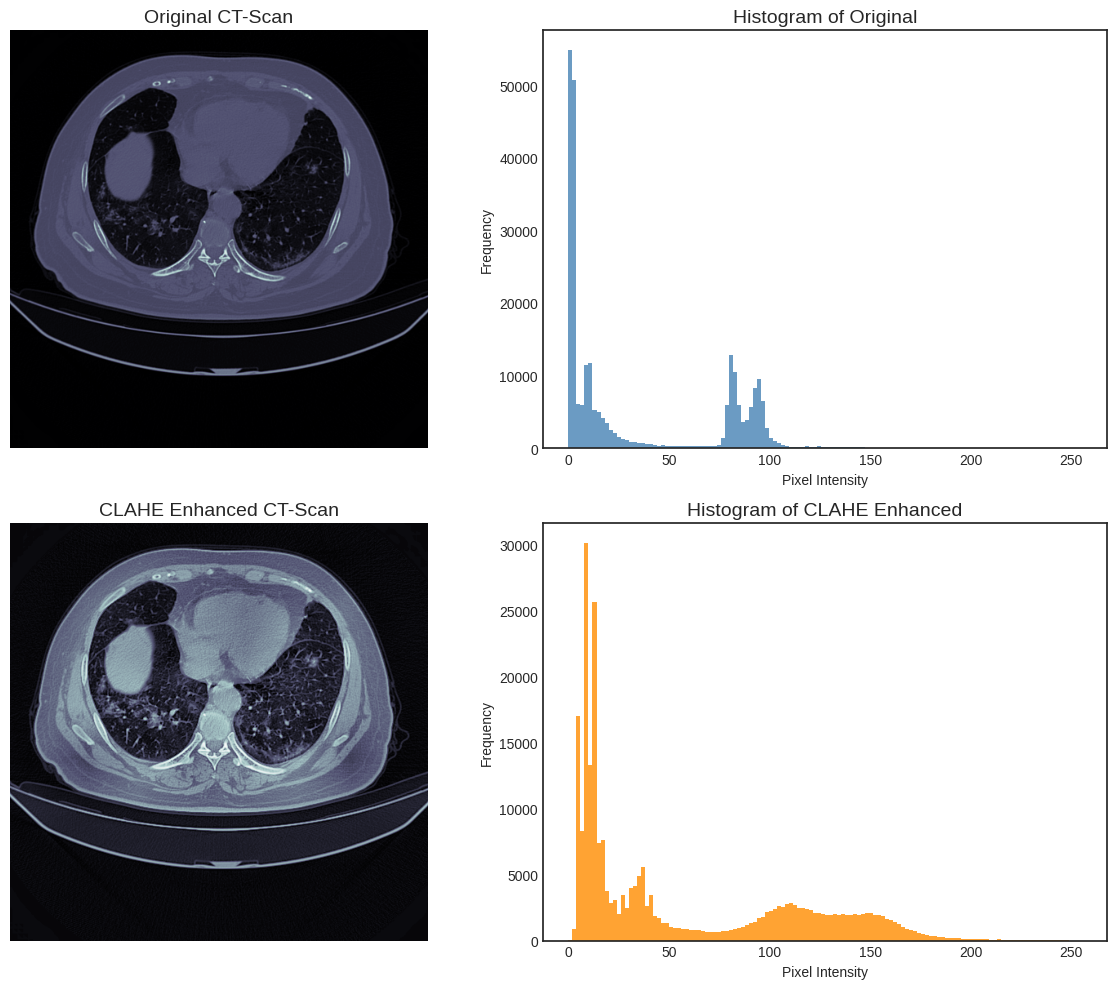}
    \caption{Comparison of original and CLAHE-enhanced CT scans with corresponding histograms}
    \label{fig:clahe}
\end{figure}

Figure \ref{fig:clahe} illustrates the effect of CLAHE on CT scans. The original CT image \textit{(top left)} and its corresponding histogram \textit{(top right)} show that most pixel intensities are clustered in narrow low-value ranges, limiting contrast between anatomical structures. After applying CLAHE \textit{(bottom left)}, the enhanced image exhibits improved visibility of lung details, while the corresponding histogram \textit{(bottom right)} demonstrates a more balanced distribution of pixel intensities across the full dynamic range. This redistribution enhances local contrast and highlights subtle structural variations that are less discernible in the original scan.

\begin{figure}[h]
    \centering
    \includegraphics[width=.75\linewidth]{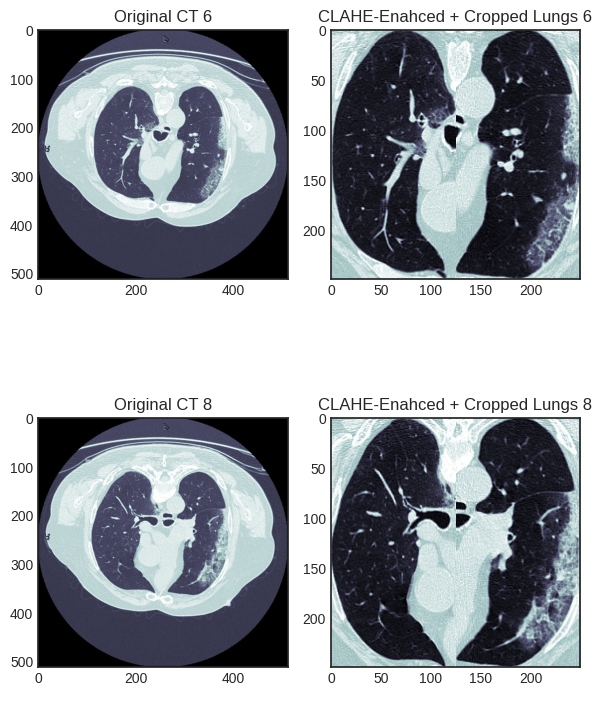}
    \caption{CT Scans and Cropped Lung Regions of COVID-19 CT segmentation dataset}
    \label{fig:cropped}
\end{figure}

The second step is a meticulous region-of-interest (ROI) extraction and cropping process. This process is specifically designed to isolate the lung regions from the rest of the CT scan. It operates by first creating a binary mask of the lungs and then using contour finding algorithms to identify the two largest contours, representing the left and right lungs. Bounding boxes are then drawn around each lung. A key design choice is to crop each lung region separately and then resize both to a consistent size of 125x250 pixels before horizontally concatenating them into a single 250x250 pixel image. As shown in figure \ref{fig:cropped}, this approach focuses the model exclusively on the diagnostically relevant pulmonary regions, removing extraneous information such as ribs, the heart, or background noise. By providing the model with a consistent, focused input, the feature learning process becomes more efficient and effective, which is a strong contributing factor to the final high performance.

\begin{figure*}[h]
    \centering
    \includegraphics[width=0.9\linewidth]{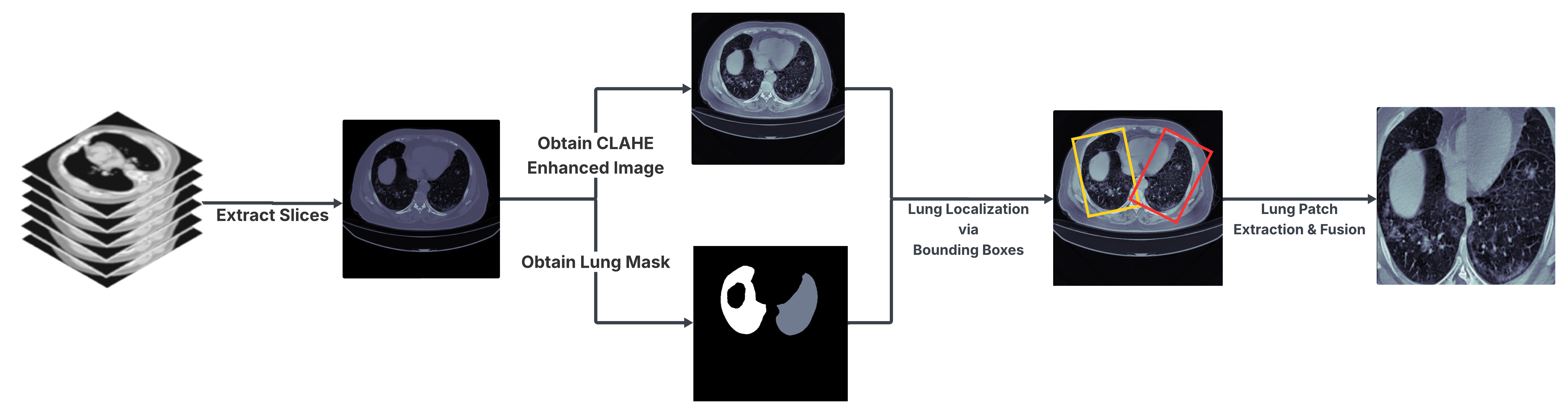}
    \caption{Pre-processing steps done for the CT scans}
    \label{fig:preprocessing}
\end{figure*}

\subsection{Data Augmentation for Dataset Expansion and Class Balancing}
The raw data, even when combined, presents a class imbalance, with 1,724 COVID-19 cases and 885 non-COVID cases. A comprehensive data augmentation strategy was implemented to address this imbalance and to significantly increase the overall size of the training dataset. The augmentation pipeline applies a variety of transformations, including rotation, horizontal and vertical flipping, shifting, gamma correction, and the addition of slight noise. These transformations artificially increase the diversity of the training data, helping the model learn features that are invariant to minor variations in image orientation or quality. The augmentation successfully balanced the dataset, resulting in a training set with 2,500 samples for each class, for a total of 5,000 images. This balancing obviated the need for a specialized loss function like Focal Loss, which was initially considered in the project's planning stages. The balanced dataset simplifies the training process, allowing for the use of a standard binary cross-entropy loss with equal class weights, which leads to a more robust and reliable model, as evidenced by the high F1-score.

\begin{figure}[h]
    \centering
    \includegraphics[width=0.9\linewidth]{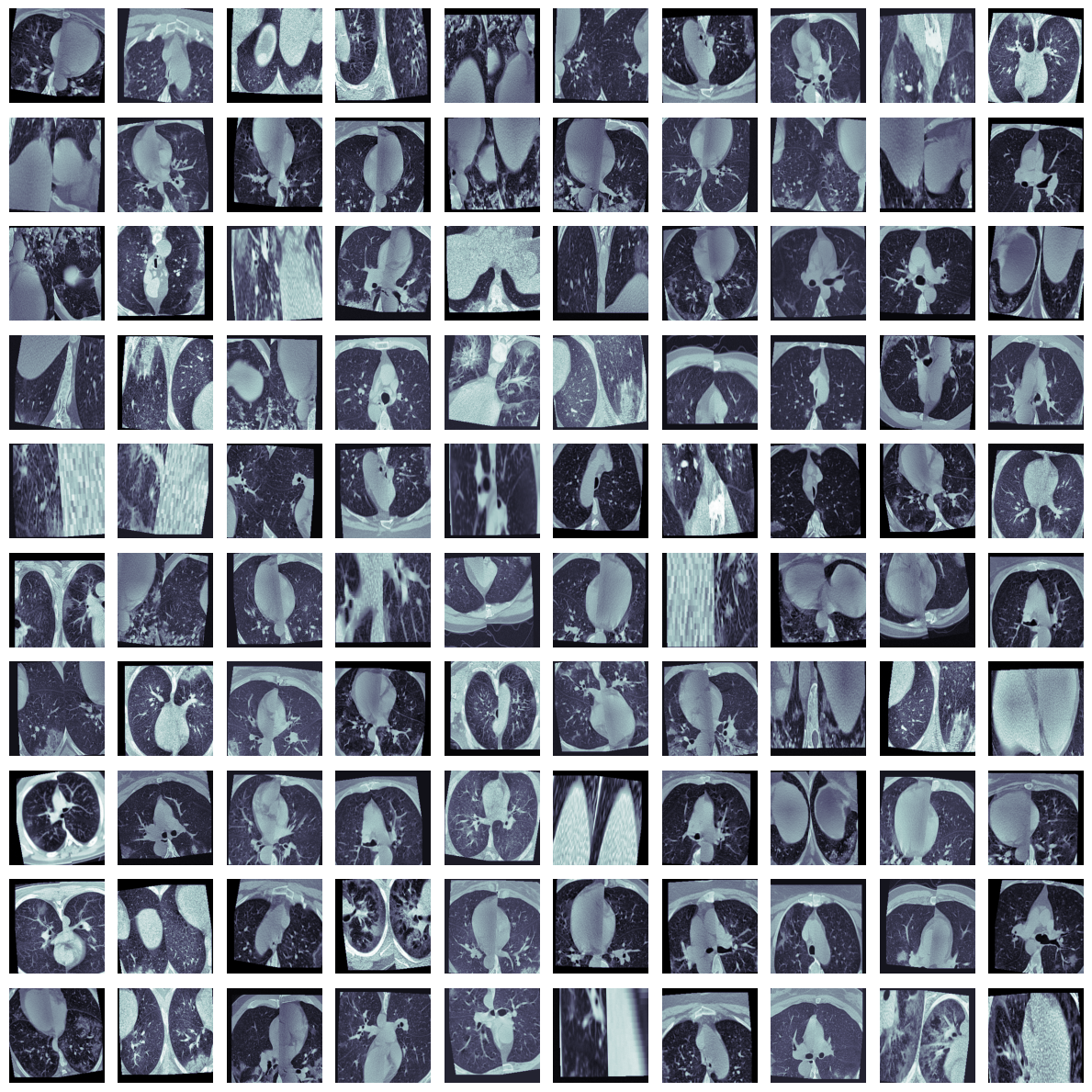}
    \caption{Generated CT scans from data augmentation}
    \label{fig:placeholder}
\end{figure}

\subsection{The Proposed Multi-branch ConvNeXt Architecture}
\begin{figure}[H]
    \centering
    \includegraphics[width=.95\linewidth]{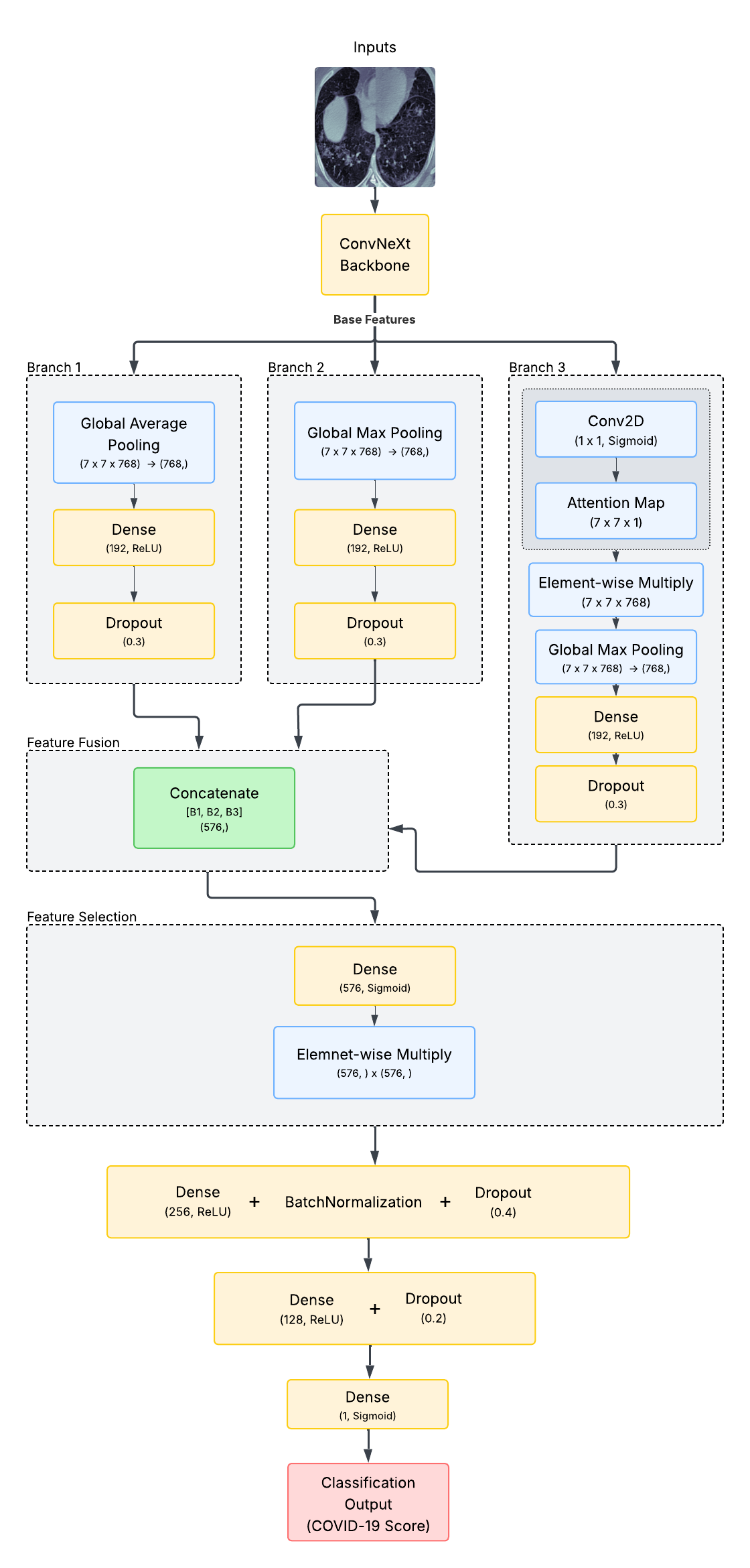}
    \caption{Multi-branch ConvNeXt Architecture}
    \label{fig:architecture}
\end{figure}

The proposed model builds on the ConvNeXtSmall architecture (Figure \ref{fig:architecture}), a high-performing convolutional model pre-trained on the ImageNet dataset. To enhance its feature extraction capabilities for the specific task of medical image diagnosis, a novel multi-branch structure was designed and implemented. This architecture processes the feature maps from the base ConvNeXt model through three parallel, distinct pathways before fusing them for final classification.

\begin{enumerate}
    \item \textbf{Global Average Pooling Branch}: This branch uses a global average pooling operation to condense the feature maps into a single vector, capturing the overall, holistic features of the image. This pathway is effective for learning general texture and context.\\
    \item \textbf{Global Max Pooling Branch}: This branch employs a global max pooling operation to identify the most salient or prominent features within the feature maps. This is particularly useful for detecting strong signals, such as large, high-intensity lesions that may indicate severe pathology.\\
    \item \textbf{Attention-weighted Pooling Branch}: This novel branch is designed to dynamically focus the model on the most diagnostically relevant regions. It learns an attention mask that is then multiplied with the base feature maps, effectively weighting the importance of each feature channel and spatial location. A global average pooling is then applied to this attention-weighted output, compelling the model to learn where to look for critical clinical markers, such as subtle ground-glass opacities.
\end{enumerate}

The outputs from these three branches are concatenated and processed through a feature selection layer, a dense layer with a sigmoid activation, which learns to weigh the importance of the combined feature vector. This enriched feature representation is then passed through a final classification head composed of dense, normalization, and dropout layers, culminating in a single-neuron output with a sigmoid activation for binary classification. This synergistic combination of different pooling strategies allows the model to gain a comprehensive understanding of the input image, capturing global patterns, strong local signals, and the most critical regions, which is a significant factor in its superior discriminative power.

\subsection{Two-Phase Training Strategy}
\begin{figure}[h]
    \centering
    \includegraphics[width=0.75\linewidth]{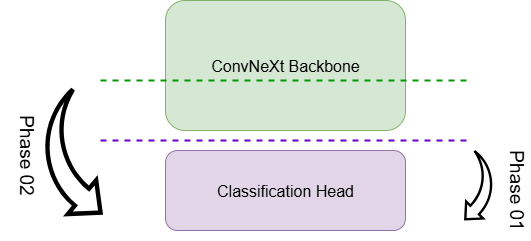}
    \caption{Two-phase training strategy}
    \label{fig:training}
\end{figure}

As shown in Figure \ref{fig:training} two-phase training strategy was employed to leverage the benefits of transfer learning while adapting the model to the specific domain of CT scans.
\\
\subsubsection{\textbf{Phase 1: Training with a Frozen Base Model}}
The initial phase focused on training only the newly added classification head while keeping the pre-trained ConvNeXt base model frozen. This phase was conducted for 12 epochs with a relatively high learning rate of $1 \times 10^{-3}$ and an adaptive optimizer. This approach allows the new layers to quickly learn how to interpret the features extracted by the pre-trained ConvNeXt model, preventing "catastrophic forgetting" of the general visual features learned from ImageNet.
\\
\subsubsection{\textbf{Phase 2: Fine-tuning with Unfrozen Layers}}
Following the initial phase, a portion of the base ConvNeXt model was unfrozen to allow for fine-tuning. Specifically, half of the base model's layers were made trainable. The model was recompiled with a much lower learning rate of $1 \times 10^{-6}$ and trained for an additional 8 epochs. This step is critical for adapting the pre-trained weights to the unique visual patterns of CT scans, enabling the model to learn the subtle, domain-specific features of pulmonary pathologies.\\

Both phases of training utilized callbacks to save the best model weights based on validation loss and AUC. Other callbacks were used to prevent overfitting and to automatically reduce the learning rate when validation performance plateaued. This structured training methodology ensures the model is both efficient to train and highly specialized for the final task.

\section{Results}
\subsection{Experimental Setup}
The combined dataset of 2,609 CT slices was split into training and validation sets using a stratified \texttt{train\_test\_split} with a 70/30 ratio, which maintained the original class distributions in both sets. After data augmentation, the final training set comprised 5,000 images, evenly split between COVID-19 and non-COVID-19 cases. The validation set consisted of 701 images after cleaning, with 486 COVID-19 cases and 215 non-COVID-19 cases. The models were trained using a batch size of 32, and an adaptive optimizer was used for both training phases.

\subsection{Performance Evaluation}
The final performance of the trained model was evaluated on the unseen validation dataset. The model achieved a high level of performance across multiple key metrics, which is a direct consequence of the robust methodology and sophisticated architecture. The final evaluation yielded a validation loss of 0.0912 and AUC of 0.9937. A detailed breakdown of the performance is provided in the classification report and the performance metrics table \ref{tab:metrics} below.

\begin{table}[h]
    \centering
    \caption{Evaluation metrics on the test set}
    \begin{tabularx}{0.6\linewidth}{X c}
        \hline
        \textbf{Metric} & \textbf{Value}\\
        \hline
        Loss      & 0.0912 \\
        AUC       & 0.9937 \\
        F1-Score  & 0.9825 \\
        Precision & 0.9835 \\
        Recall    & 0.9815 \\
        Accuracy  & 0.9757 \\
        \hline
    \end{tabularx}
    \label{tab:metrics}
\end{table}

\begin{figure}[h]
    \centering
    \includegraphics[width=0.75\linewidth]{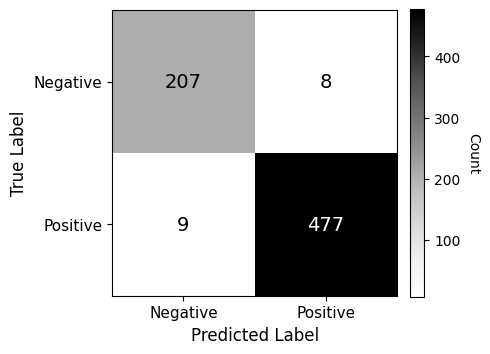}
    \caption{Confusion matrix of the classification results}
    \label{fig:cm}
\end{figure}

The confusion matrix, figure \ref{fig:cm} further illustrates the model's high accuracy, with only 8 false positives and 9 false negatives out of 701 validation images. Reported precision, recall, and F1-score are calculated from it, and this provides a transparent view of class-wise performance and ensures reproducibility of the reported metrics.

Given the inherent class imbalance in the original dataset, with a greater number of COVID-19 cases, metrics beyond simple accuracy are crucial for a comprehensive evaluation. Accuracy can be misleading because a model can achieve a high score by simply predicting the majority class, which would be a poor outcome for a clinical diagnostic tool. The AUC and F1-score provide a more reliable assessment. AUC measures the model's ability to distinguish between the two classes across all possible classification thresholds, which is a vital indicator of its overall discriminating power. The F1-score is the harmonic mean of precision and recall, providing a balanced measure that is particularly useful for imbalanced datasets, as it penalizes models that favor one class over the other. 

Therefore, during training and evaluation, our models were optimized to maximize AUC rather than raw accuracy. This choice ensures that the reported performance better reflects real-world diagnostic needs, where sensitivity (minimizing missed COVID cases) and specificity (avoiding false alarms) are both critical. The source code is publicly available online \cite{b24}
\\\\
\textbf{\textit{Key Performance Metrics Explained}}\\
The performance of a classification model is measured using a variety of metrics, each providing a different perspective on its effectiveness. These metrics are derived from the values in the confusion matrix: True Positives (TP), True Negatives (TN), False Positives (FP), and False Negatives (FN).
\begin{itemize}
    \item \textbf{Accuracy}: The ratio of correctly predicted observations to the total observations.
\[
Accuracy = \frac{TP + TN}{TP + TN + FP + FN}
\]
\item \textbf{Precision}: The ratio of correctly predicted positive observations to the total predicted positive observations. It answers the question: "\textit{Of all the cases we predicted as positive, how many were actually positive?}"
\[
Precision = \frac{TP}{TP + FP}
\]
\item \textbf{Recall}: The ratio of correctly predicted positive observations to all observations in the actual class. It answers the question: "\textit{Of all the actual positive cases, how many did we correctly identify?}" Also known as Sensitivity or True Positive Rate (TPR).
\[
Recall = \frac{TP}{TP + FN}
\]
\item \textbf{F1-Score}: The weighted average of Precision and Recall. It is a single metric that balances both concerns and is especially useful in cases with uneven class distribution.
\[
F_{1} = \frac{2 \cdot \\Precision \cdot \\Recall}{\\Precision + \\Recall}
\]
\item \textbf{Area Under the Curve (AUC)}: A measure of the model's ability to distinguish between classes. The AUC for the Receiver Operating Characteristic (ROC) curve plots the True Positive Rate (Recall) against the False Positive Rate (FPR), which is the ratio of incorrectly predicted positive observations to all actual negative observations. A higher AUC indicates a better model.
\[
FPR = \frac{FP}{FP + TN}
\]
\end{itemize}

\subsection{Comparative Analysis}
The results of the proposed model were evaluated against contemporary state-of-the-art benchmarks reported in the literature. Importantly, all comparisons were performed on the same datasets, namely the \textbf{COVID-19 CT Lung and Infection Segmentation Dataset}\cite{b18} and the \textbf{MedSeg Covid Dataset 2}\cite{b5}, ensuring fairness and consistency in evaluation. As shown in the table \ref{tab:comparison} below, the \textbf{multi-branch ConvNeXt} model achieves competitive performance, demonstrating its effectiveness in comparison to the best-performing models in the field.

\begin{table}[h!]
\centering
\caption{Performance comparison of different methods}
\begin{tabularx}{\columnwidth}{lXXXX}
\hline
\textbf{Methods} & \textbf{Acc} & \textbf{Sen} & \textbf{Spe} & \textbf{AUC} \\
\hline
CNN 8-layers \cite{b23}   & 0.7467     & 0.8 & 0.70 & 0.78\\
InceptionV3 \cite{b23}   & 0.8267 & 0.88 & 0.78 & 0.82 \\
Efficient-Net \cite{b23} & 0.9067 & 0.91 & 0.85 & 0.93 \\
ResNet+SE \cite{b21}     & 0.8707 & 0.9322 & 0.8030 & 0.9557 \\
ResNet \cite{b19}        & 0.8890 & 0.9253 & 0.8439 & 0.9649 \\
ResNet+CBAM \cite{b20}   & 0.9162 & 0.8808 & 0.9552 & 0.9784  \\
MTL \cite{b23}           & 0.9467  & 0.96 & 0.92 & 0.97 \\
MA-Net \cite{b22}         & 0.9588 & 0.9512 & 0.9672 & 0.9885 \\
\textbf{Ours (\textit{MB-ConvNeXt})}                     & \textbf{0.9757} & \textbf{0.9815} & \textbf{0.9835} & \textbf{0.9937}\\
\hline
\end{tabularx}
\label{tab:comparison}
\end{table}

\subsection{Qualitative Analysis and Visual Insights}

\begin{figure}[h]
    \centering
    \includegraphics[width=.8\linewidth]{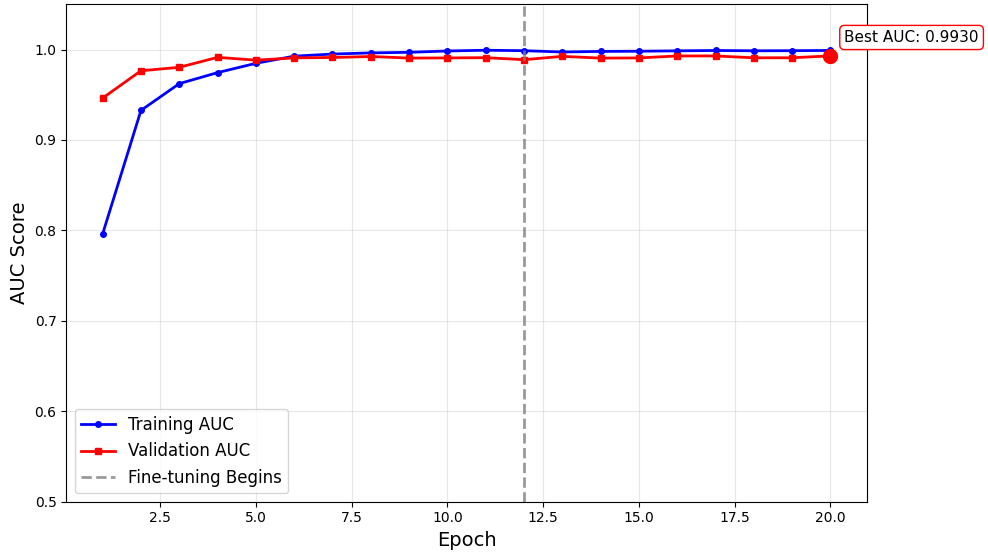}
    \caption{AUC evolution plot}
    \label{fig:auc-evol}
\end{figure}

Beyond the quantitative metrics, a qualitative analysis was performed to gain a deeper understanding of the model's behavior. The AUC evolution plot (Figure \ref{fig:auc-evol}) shows a steep increase during the initial training phase, indicating that the classification head rapidly adapted to the pre-trained ConvNeXt features. After fine-tuning began at epoch 12, both training and validation curves stabilized, with validation AUC peaking at 0.9937. The close alignment between the two curves suggests strong generalization and minimal overfitting.

\begin{figure}[h]
    \centering
    \includegraphics[width=.75\linewidth]{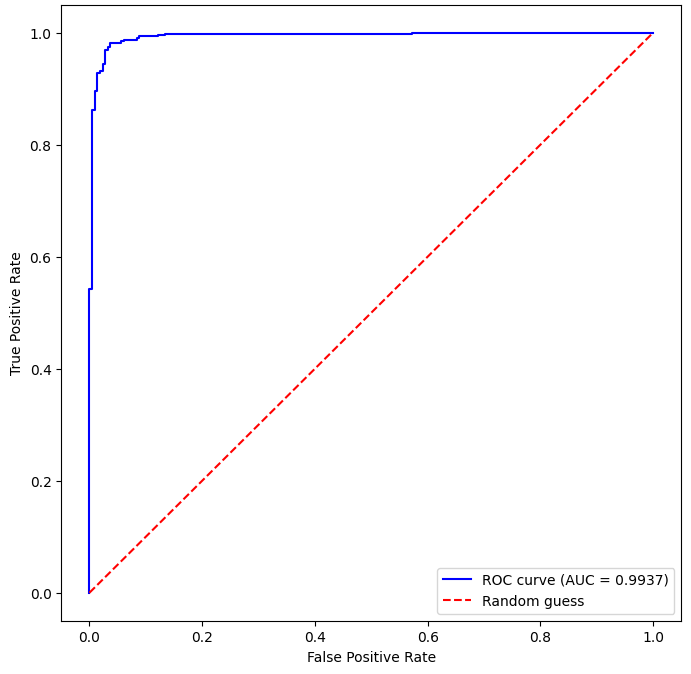}
    \caption{ROC curve}
    \label{fig:auc}
\end{figure}

The final ROC curve plot (Figure \ref{fig:auc} also confirms the model's excellent performance, with a curve that closely follows the top-left corner of the graph, indicating a strong ability to distinguish between the two classes.

\begin{figure}[h]
    \centering
    \includegraphics[width=0.8\linewidth]{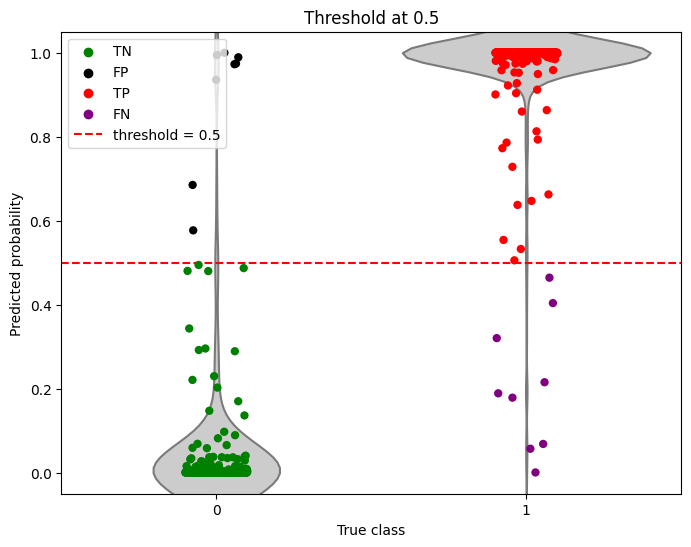}
    \caption{Predicted probability distributions}
    \label{fig:violin}
\end{figure}

An analysis of the predicted probability distributions using a violin plot (Figure \ref{fig:violin} reveals that the model is highly confident in its predictions for both classes, with the distributions for COVID and non-COVID cases showing clear separation. This visualization confirms the model's robust discriminative capabilities.

\section{Discussion}
The results demonstrate that the proposed multi-branch ConvNeXt model, in combination with a detailed preprocessing and training pipeline, is a powerful tool for COVID-19 diagnosis from CT scans. The high precision (0.9835), recall (0.9815), and F1-score (0.9825) for the COVID-19 class are particularly significant for a diagnostic application, as they indicate that the model is both highly effective at identifying true positive cases and robust in avoiding false positives. The high ROC-AUC of 0.9937 confirms the model's excellent ability to discriminate between the two classes across various classification thresholds.

The success of this approach is attributed to several key methodological choices. First, the combination of data from two distinct sources allowed for a larger training set, which is fundamental for learning robust features. Second, the meticulous preprocessing steps, CLAHE enhancement and the two-lung cropping and concatenation, acted as a form of a manual attention mechanism, directing the model's focus to the diagnostically relevant regions and thereby simplifying the learning task. Third, the data augmentation pipeline effectively solved the class imbalance problem, which allowed for a more straightforward training process with binary cross-entropy loss and equal class weights. This avoided the complexities of a specialized loss function and ensured the model learned a balanced representation of both classes.

Finally, the novel multi-branch architecture itself is a core factor in the model's performance. By combining features from global average, global max, and a learned attention-weighted pooling branch, the model is able to synthesize a more comprehensive understanding of the input image. This design allows it to capture a wider array of visual cues, from general texture to the most salient features and the most critical regions, which are all important for an accurate diagnosis. The two-phase training strategy, which first allowed the classification head to learn and then fine-tuned the base model, was also crucial for adapting the pre-trained weights to the unique visual patterns of CT scans without over-fitting.

Despite the high performance, certain limitations should be noted. The dataset, while sufficient for a robust proof-of-concept, is still relatively small compared to what might be available in a large-scale clinical setting. The model's generalizability could be further improved by training on a larger, more diverse dataset from multiple hospitals and with varied scanner parameters. Future work could also involve exploring alternative architectures, such as a full Vision Transformer, or a hybrid model that combines convolutional and transformer layers, to see if they can achieve even higher performance.

\section{Conclusion}
This research successfully demonstrates a multi-faceted approach to automated COVID-19 diagnosis from CT scans. The proposed methodology, which includes a novel multi-branch ConvNeXt architecture and a rigorous end-to-end training pipeline, significantly improves upon the performance of early deep learning models and achieves results that are competitive with contemporary state-of-the-art benchmarks.

The project’s success is a testament to the power of modern deep learning techniques when applied systematically and with careful consideration of domain-specific challenges. The use of data augmentation to balance the training set, the meticulous preprocessing to focus the model on the region of interest, and the multi-branch architecture to capture a comprehensive feature representation were all key to achieving the final high performance. The culminating model, with an ROC-AUC of 0.9937 and an F1-score of 0.9825, represents a significant step forward in developing robust and clinically useful AI tools for medical diagnostics. The principles of this multi-faceted approach, including advanced preprocessing, architectural enhancements, and a structured training regimen, are broadly applicable to a wide range of medical image classification challenges beyond COVID-19. The findings from this research provide a valuable contribution to the field and offer a clear path for future work aimed at translating these models into practical clinical applications.

\end{document}